\documentclass[letterpaper, 10 pt, journal, twoside]{IEEEtran}
\ifCLASSINFOpdf
\else
\fi

\usepackage{amsmath} 
\usepackage{amssymb}  
\usepackage{svg}
\usepackage{caption}
\usepackage{multirow}

\usepackage{amsfonts}
\usepackage{array}
\usepackage[caption=false,font=normalsize,labelfont=sf,textfont=sf]{subfig}
\usepackage{textcomp}
\usepackage{stfloats}
\usepackage{url}
\usepackage{verbatim}
\usepackage{graphicx}
\usepackage{cite}
\usepackage{soul}

\usepackage{graphics}
\usepackage[linesnumbered,ruled,vlined]{algorithm2e}
\usepackage{makecell}
\usepackage{multirow}
\usepackage{rotating}
\usepackage{mathtools}

\SetKwInput{KwInput}{Input}                
\SetKwInput{KwOutput}{Output}  
\SetKwRepeat{Do}{do}{while}%

\begin{document}
%
\title{D2S: Representing sparse descriptors and 3D coordinates for camera relocalization}
%
%
%

\author{Bach-Thuan Bui$^{1}$, Huy-Hoang Bui$^{1}$, Dinh-Tuan Tran$^{2}$, and Joo-Ho Lee$^{2}$%
\thanks{Manuscript received: June 21, 2024; Revised September 20, 2024; Accepted October 18, 2024.}
\thanks{This paper was recommended for publication by Editor Vasseur Pascal upon evaluation of the Associate Editor and Reviewers' comments.}
\thanks{$^{1}$B. Bui and H. Bui are with Graduate School of Information Science and Engineering, Ritsumeikan University, Japan.
        {\tt\footnotesize thuan.aislab@gmail.com}}%
\thanks{$^{2} $D. Tran and J. Lee are with School of Information Science and Engineering, Ritsumeikan University, Japan}%
\thanks{The source code, trained models, dataset, and demo videos are available at the following link: https://thpjp.github.io/d2s}
\thanks{Digital Object Identifier (DOI): see top of this page.}
}
%
%

\markboth{IEEE Robotics and Automation Letters. Preprint Version. Accepted October, 2024}
{Bui \MakeLowercase{\textit{et al.}}: D2S}

%



\maketitle

\begin{abstract}

State-of-the-art visual localization methods mostly rely on complex procedures to match local descriptors and 3D point clouds. However, these procedures can incur significant costs in terms of inference, storage, and updates over time. In this study, we propose a direct learning-based approach that utilizes a simple network named D2S to represent complex local descriptors and their scene coordinates. Our method is characterized by its simplicity and cost-effectiveness. It solely leverages a single RGB image for localization during the testing phase and only requires a lightweight model to encode a complex sparse scene. The proposed D2S employs a combination of a simple loss function and graph attention to selectively focus on robust descriptors while disregarding areas such as clouds, trees, and several dynamic objects. This selective attention enables D2S to effectively perform a binary-semantic classification for sparse descriptors. Additionally, we propose a simple outdoor dataset to evaluate the capabilities of visual localization methods in scene-specific generalization and self-updating from unlabeled observations. Our approach outperforms the previous regression-based methods in both indoor and outdoor environments. It demonstrates the ability to generalize beyond training data, including scenarios involving transitions from day to night and adapting to domain shifts.

\end{abstract}


\section{Introduction}

Visual re-localization plays a crucial role in numerous applications including robotics and computer vision. The majority of methods rely on sparse reconstruction models, and local feature matching \cite{detone2018superpoint, sarlin2019coarse}. The 3D reconstructed model, commonly built using Structure from Motion (SfM), stores 3D points cloud map along with local and global visual descriptors \cite{detone2018superpoint, lowe2004distinctive, arandjelovic2016netvlad, sarlin2019coarse}. While showing superior performance in handling appearance changes introduced by different devices and viewpoints, these methods suffer from several drawbacks. First, point clouds and visual descriptors need to be stored explicitly, which hinders their practicality to large-scale scenes due to storage requirements. Second, the involvement of multiple components such as feature extraction and feature matching make these approaches computation-extensive which is not suitable for execution on consumer or edge devices. 

Conversely, end-to-end learning of the implied 2D--3D correspondences offers a better trade-off between accuracy and computational efficiency \cite{brachmann2017dsac, li2020hierarchical, brachmann2021visual}. These approaches, commonly referred to as scene coordinate regression (SCR), directly regressing scene coordinates from corresponding 2D pixels of the input image. Thus, SCRs naturally establish the 2D--3D matches without the need for the 3D models but it is worth mentioning that they are prone to overfitting if the training data does not adequately represent the environment \cite{li2018full}. Although previous SCR-based approaches have demonstrated high accuracy for re-localization in small-scale and stationary environments \cite{brachmann2017dsac,brachmann2018learning,brachmann2021visual}, their performance are not on par with feature matching (FM)-based methods in environments that contain variations over a long span of time \cite{do2022learning} and high domain shift scenarios.

\begin{figure}
    \centering
    \includegraphics[width=0.45\textwidth]{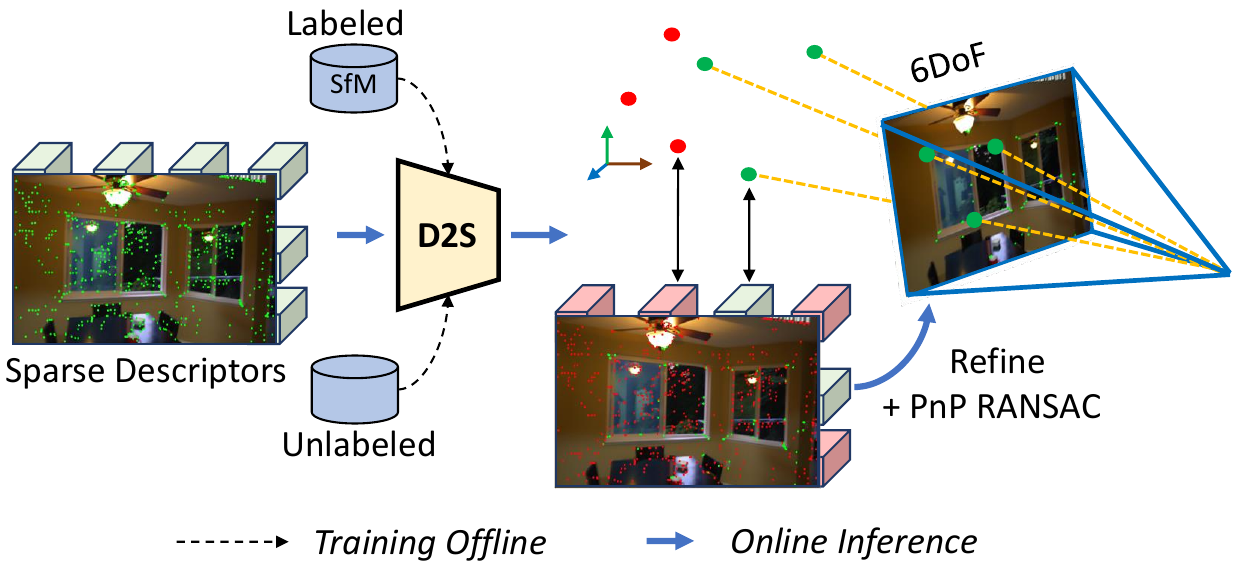}
    \caption{\textbf{D2S learning and inference pipeline}. Our D2S learns from labeled descriptors and is able to self-update with unlabeled ones. At the test time, D2S receives a set of local descriptors and generates robust 3D coordinates for highly accurate re-localization.}
    \label{fig1}
\end{figure}

In this study, we introduce a new visual localization approach named D2S that leverages keypoint descriptors to address the aforementioned issues of previous SCR methods. Instead of learning directly from RGB images, we initiate the regression network from sparse salient descriptors. This allows the network to learn the mapping between keypoint descriptors and their corresponding scene coordinates. By exploiting the self-attention mechanism, we also enable D2S to learn the global context by communicating between features. As a result, D2S inherits the robustness of salient features, making it resilient against drastic changes in the environment. In addition, we explore an aspect that previous SCR research \cite{brachmann2021visual, brachmann2023accelerated} lacks, which is the ability of the model to utilize new, unlabeled data from various devices. We propose a simple yet efficient self-supervised technique for updating the re-localizer with unlabeled observations, which further enhances D2S's ability to accommodate dynamic environments. In Fig. \ref{fig1}, we present a summary of the proposed pipeline. Our contributions can be listed as follows:

\begin{itemize}
    \item We introduce D2S (Descriptors to Scenes), a novel lightweight network for representing sparse descriptors and 3D coordinates, enabling real-time, high-precision re-localization. 
    \item The proposed D2S combines graph attention with a simple loss function to identify reliable descriptors, enabling a strong focus on robust descriptors and improving re-localization performance. 
    \item We show that D2S can seamlessly self-update with new observations without requiring camera poses, parameters, or ground truth scene coordinates. To this end, we introduce a small challenge dataset to evaluate the method in scenarios of high domain shift, sparse training data, and self-updating with unlabeled data.
    \item We present a comprehensive evaluation of the proposed D2S across three indoor and outdoor datasets, demonstrating that our method outperforms the state of the art.
\end{itemize}

\section{Related Work} \label{related_work}
\subsection{Image Retrieval and Absolute Pose Regression}

Image-based visual re-localization methods\cite{schindler2007city} were arguably the simplest solution, which constructing maps just by storing a set of images and their camera pose in a database. Given a query image, we seek for the most similar images in the database by comparing their global descriptor \cite{arandjelovic2016netvlad, torii201524}. With the matched images representing a location in a scene that is close to the query image \cite{torii201524, arandjelovic2016netvlad}, the pose of the query image can be derived from the top retrieved images pose. Despite demonstrating scalability in large environments, these approaches have limited accuracy in determining camera poses. Thus they are often used as initialization for later pose refinements.

6DoF absolute camera pose regression approaches (APR) have been studied extensively to overcome the drawbacks of image retrieval. Notably, PoseNet\cite{kendall2015posenet} was first proposed, along with subsequent developments \cite{kendall2015posenet, bach2022featloc, chen2022dfnet}. APR learns to directly encode the 6DoF camera pose from the image, resulting in a lightweight system where memory demand and estimation time are independent of the reference data size. However, theoretical conclusions \cite{sattler2019understanding} suggest that APR behaviors are more closely related to image retrieval than methods that estimate poses by 3D geometry. Thus APR could hardly outperform a handcraft retrieval baseline.

\begin{figure*}[t]
    \centering
    \includegraphics[width=460pt]{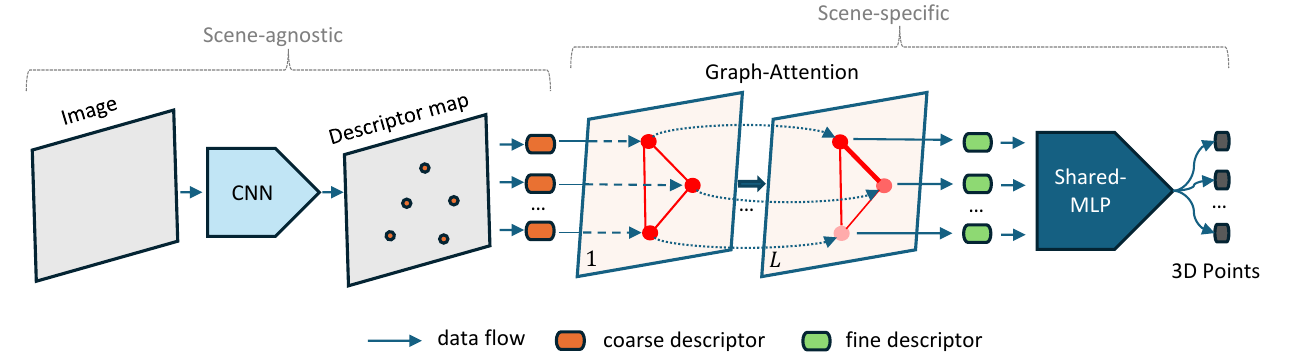}
    \caption{\textbf{D2S architecture}. The proposed D2S pipeline enables direct learning to represent 3D coordinates from sparse keypoint descriptors. The method comprises a \textit{scene-agnostic} feature extractor that extracts sparse descriptors from images, followed by a learnable \textit{scene-specific} module aimed at regressing their 3D coordinates and reliability scores by leveraging multi-layer graph attention, and a shared MLP.}
    \label{fig:network_architecture}
\end{figure*}

\subsection{Sparse Feature Matching}

Features matching-based (FM) re-localizers compute the camera pose from 2D-3D correspondences between 2D pixels and 3D scene maps, constructed using Structure-from-Motion (SfM) \cite{schonberger2016structure, sarlin2019coarse}. Early FM-based solutions match local descriptors in 2D images to 3D points in the SfM model \cite{svarm2016city, toft2018semantic, sattler2016efficient, liu2017efficient}. Despite achieving accurate pose estimations, exhaustive matching creates a substantial computational bottleneck. Additionally, as SfM models grow in size, direct matching becomes ambiguous and difficult to establish under significant appearance changes. Leveraging the robustness of image retrieval in scalability and handling appearance changes, \cite{sarlin2019coarse, sarlin2020superglue} proposed a hierarchical re-localization approach that uses image retrieval as an initialization step before performing 2D-3D direct matching. Instead of performing complex 2D-3D matching, \cite{pietrantoni2023segloc} proposed deep feature alignment, reducing system complexity while preserving accurate estimation.

\subsection{Scene Coordinate Regression}

Compared to the FM-based pipeline, scene coordinate regression (SCR) approaches are more straightforward. By employing a single regression network, 3D coordinates can be directly extracted from 2D pixels \cite{ brachmann2017dsac, brachmann2018learning, li2020hierarchical, brachmann2021visual, brachmann2023accelerated}. These correspondences then serve as the input for RANSAC-based pose optimization. SCR methods have demonstrated superior accuracy in camera pose estimation in small-scale and stationary environments. However, scaling these methods to larger, dynamic environments remains challenging. They struggle with adapting to novel viewpoints and handling the increased complexity that comes with larger-scale environments \cite{do2022learning}. Our proposed model, adopting similar idea to SCR but aim to address the aforementioned problem. By utilizing the robustness of local keypoint descriptors, our model seeks to enhance performance in high viewpoint changes and dynamic scenarios.

\subsection{Self-Supervised Updating with Unlabeled Data}

Visual SLAM systems \cite{sumikura2019openvslam} allow continuous map updates, but most visual re-localization techniques, including FM \cite{sarlin2019coarse, yang2022scenesqueezer} and SCR-based methods \cite{ brachmann2018learning, li2020hierarchical, brachmann2021visual, brachmann2023accelerated}, lack such adaptability, requiring reconstruction and labeling for updates. APR-based approaches have pioneered self-supervised updates with unlabeled data, utilizing sequential frames \cite{brahmbhatt2018geometry} or feature matching \cite{chen2022dfnet}. ACE0 \cite{brachmann2024scene} represents the first SCR-based method to introduce this capability but remains limited under highly dynamic conditions \cite{do2022learning}. In this study, we demonstrate that the proposed sparse regression-based D2S adapts naturally to such conditions through self-supervision without needing reconstruction or labeling.



\section{Proposed Approach} \label{proposed_method}

Given a set of reference images $\{\mathbf{I}^{i}\}_{i=1}^{n}$ and its reconstructed SfM model $\mathcal{E}$,  we aim to develop a sparse regression module, which can encode the entire environment $\mathcal{E}$ using a compact function $\mathfrak{F}(.)$, where $\mathfrak{F}$ is a deep neural network. The proposed function $\mathfrak{F}(.)$ inputs a set of local descriptors $\{\mathbf{d}^{i}\}^{k}$ extracted from $\mathbf{I}^{i}$ and outputs its corresponding 3D global cloud coordinates $\{\mathbf{w}^{i} \in \mathbb{R}^{3}\}^{k}$. The ultimate goal of the proposed module is to perform \textit{visual localization}, a task of estimating camera pose $\mathbf{T} \in \mathbb{R}^{4\times4}$ of the query image $\mathbf{I}_{q}$. We illustrate the proposed architecture in Fig. \ref{fig:network_architecture}.

\subsection{Representing Sparse Descriptors and 3D Coordinates} \label{simple_function}
Inspired by the universal continuous set function approximator \cite{bui2022fast}, we first propose a simple element set learning function $\mathfrak{F}(.)$. It receives a set of descriptors $\{\mathbf{d}^{i}\}_{i=0}^{k}$ as input and outputs corresponding scene coordinates $\{\mathbf{w}^{i}\}_{i=0}^{k}$. Overall, the proposed function can be described as follows:

\begin{equation}
\begin{aligned}
\mathfrak{F}\left(\{\mathbf{d}^{i}\}_{i=1}^{k}\right) & = \left\{f(\mathbf{d}^{i})\right\}_{i=1}^{k}\\
& = \{\mathbf{w}^{i}\}_{i=1}^{k},
\label{ori_equation}
\end{aligned}
\end{equation}

where $\mathfrak{F}:\mathbb{R}^{k \times D} \to \mathbb{R}^{k \times 4}$ and $f:\mathbb{R}^{D} \to \mathbb{R}^{4}$. $k$ is the number of descriptors, and $D$ is descriptor dimension. The function $f(.)$ is a shared nonlinear function. It receives a descriptor vector $\mathbf{d} \in \mathbb{R}^{D}$ and outputs a scene coordinate vector $\mathbf{w} = (x,y,z,p)^T$. We introduce an additional dimension $p$ for the scene coordinate, which represents the reliability probability for localization of the input descriptor $\mathbf{d}$. To ensure that the range of the reliability probability output lies within $[0,1]$ while enabling function $f(.)$ to produce $p\in \mathbb{R}$, we compute the final reliability prediction using the following equation: 

\begin{equation}
\begin{aligned}
\hat{z} = \frac{1}{1+|\beta p|} \in (0,1],
\end{aligned}
\label{reliable_equation}
\end{equation}

where $\beta$ is a scale factor chosen to make the expected value of reliability prediction $\hat{z}$ easy to reach a small value when the input descriptors belong to high uncertainty regions.

The proposed module is theoretically simple. As illustrated in Eq. \ref{ori_equation}, it requires only a non-linear function $f$ to compute the scene coordinates of given descriptors.  In practice, we approximate $f$ using a multi-layer perceptron (MLP) network, which is similar to \cite{brachmann2023accelerated}.

\begin{figure}[t]
    \centering
    \hspace*{-0.2cm}
    \includegraphics[width=0.45\textwidth]{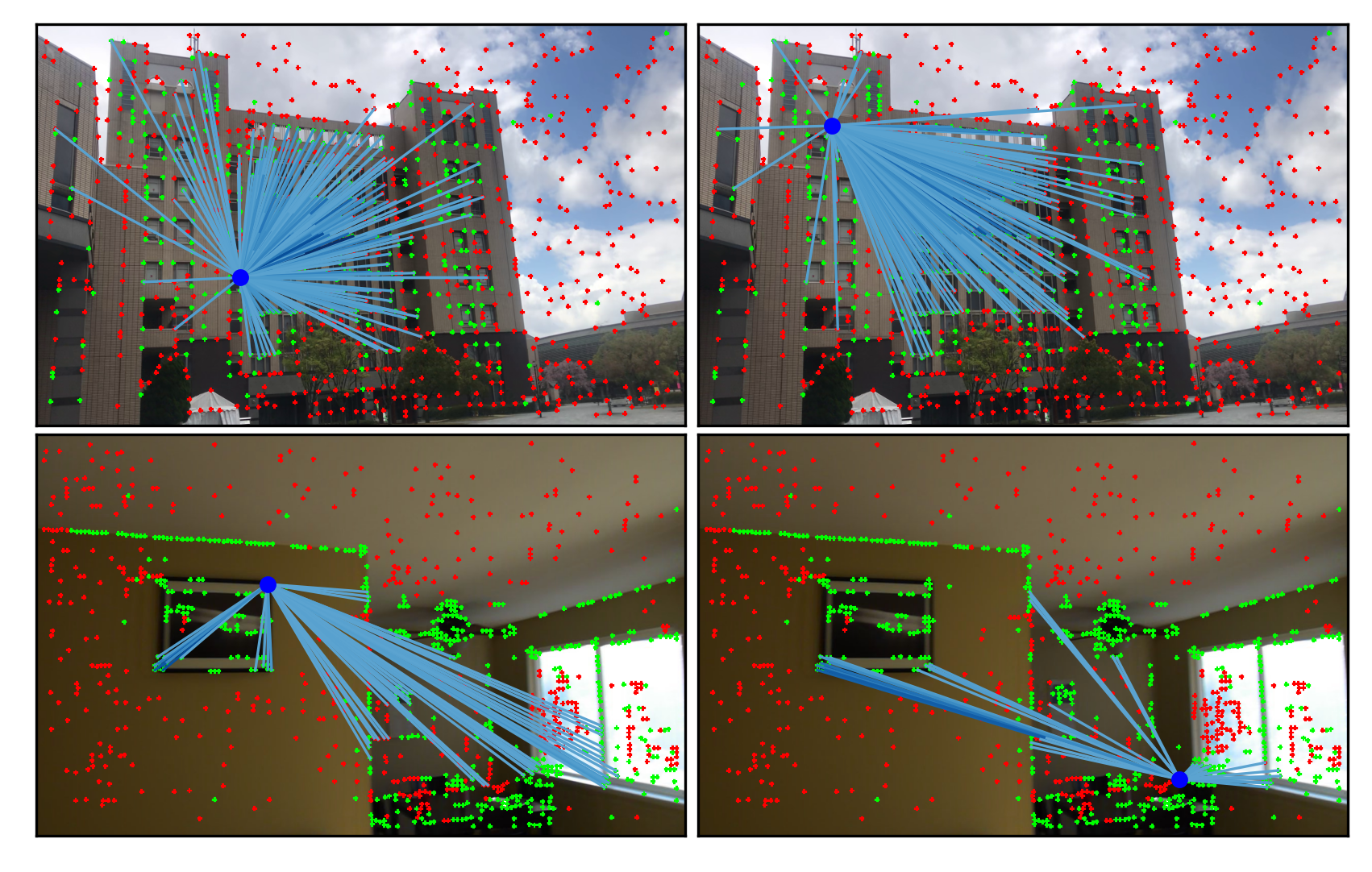}
    \caption{\textbf{Self-Attention Results}. The visualization shows the attention scores $\alpha_{ij}$ at graph layer 5. It demonstrates that D2S has succeeded in learning to focus on \textcolor{green}{reliable features} while disregarding the \textcolor{red}{uncertain features}.} 
    \label{fig_attention_1}
\end{figure}

\subsection{Optimizing with Graph Attention} \label{graph_attension}
Solely relying on independent descriptors to regress scene coordinates may lead to an insufficient understanding of the global context and the distinctiveness of feature clusters. To address this issue, we draw inspiration from recent works on feature matching \cite{sarlin2020superglue}, which leverage attention aggregation \cite{vaswani2017attention} to learn both local and global attributes of sparse features. We employ a simplified version for the scenes regression task, which can be described as follows:

\begin{equation}
\begin{aligned}
\mathfrak{F}\left(\{\mathbf{d}^{i}\}_{i=1}^{k}\right) & = f \circ  \mathcal{A}^{l}\circ... \mathcal{A}^{1}\left(\{\mathbf{d}^{i} \}_{i=1}^{k}\right)  \\
& = \{\mathbf{w}^{i}\}_{i=1}^{k},
\end{aligned}
\label{ori_grap_re}
\end{equation}

where $\mathcal{A}:\mathbb{R}^{k \times D} \to \mathbb{R}^{k \times D}$, and $\mathcal{A}^{l}\circ... \mathcal{A}^{1}(.)$ represents a nested residual attention update for each descriptor $\mathbf{d}^{i}$ across $l$ graph layers. Given a coarse local descriptor $\mathbf{d}^{i}$, $\mathcal{A}^{l}(.)$ is used to compute its \textit{fine regression 
descriptor} $\prescript{(l)}{}{\mathbf{d}^{i}}$  by letting it communicate with each other across $l$ attention layers. This implicitly incorporates visual relationships and contextual cues with other descriptors in the same image, enabling it to extract robust features automatically through an end-to-end learning pipeline from scene coordinates. In addition, the proposed reliability-aware detection, as described in Eq. \ref{reliable_equation}, can also encourage our attention module to favor features from reliable areas (e.g. buildings, stable objects) and suppress those from unreliable regions (e.g. trees, sky, human).

As $\prescript{(l)}{}{\mathbf{d}^{i}}$ is the representation for the output \textit{regression descriptor} of element $i$ at layer $l$, Eq. \ref{ori_grap_re} can be simplified as follows: 

\begin{equation}
\begin{aligned}
\mathfrak{F}(\{\mathbf{d}^{i}\}_{i=1}^{k}) & = \left\{f(\prescript{(l)}{}{\mathbf{d}^{i}}) \right\}_{i=1}^{k},
\end{aligned}
\end{equation}

which now has a similar form to the original Eq. \ref{ori_equation}, where the final regression layer $f$ is simply a shared MLP. 

Similar to a previous study \cite{sarlin2020superglue}, we also consider the attentional module as a multi-layer graph neural network. However, we only leverage the self-edge, which is based on self-attention \cite{vaswani2017attention}, to connect the descriptor $i$ to all others in the same image. The message passing formulation for element $i$ at the layer $(l+1)$ is described as

\begin{equation}
    \prescript{(l+1)}{}{\mathbf{d}^{i}}  = \prescript{(l)}{}{\mathbf{d}^{i}} + MLP\left(\left[\prescript{(l)}{}{\mathbf{d}^{i}} || \mathbf{m}_{\mathfrak{I}\rightarrow i}\right]\right),
\end{equation}

where $\mathbf{m}_{\mathfrak{I}\rightarrow i}$ denotes the aggregated message result from all descriptors $\{\prescript{(l)}{}{\mathbf{d}^{i}}\}_{i=1}^{k}$, with  $\mathfrak{I}=\{1,...,k\}$ indicating the set of descriptor indices, and $[.||.]$ denoting concatenation. This module consists of $L$ chained layers with different parameters. Therefore, the message $\mathbf{m}_{\mathfrak{I}\rightarrow i}$ is different in each layer. 

The self-attention mechanism performs an interaction to map the query vector $\mathbf{q}_{i}$ against a set of key vectors $\{\mathbf{k}_{j}\}_{j \in \mathfrak{I}}$, associated with candidate descriptors, to compute the attention score $\alpha_{ij}=$ Softmax$_{j}(\mathbf{q}_{i}^{T}\mathbf{k}_{j})$. It then presents the best-matched descriptors with their value vectors $\{\mathbf{v}_{j}\}_{j \in \mathfrak{I}}$ represented by a scored average of values, which in practice is the computation of message $\mathbf{m}_{\mathfrak{I}\rightarrow i}$:

\begin{equation}
\begin{aligned}
\mathbf{m}_{\mathfrak{I}\rightarrow i} = \sum_{j:(i,j)\in\mathfrak{I}}{\alpha_{ij}\mathbf{v}_{j}}.
\end{aligned}
\end{equation}

The query, key, and value are derived from linear projections between the descriptor $\prescript{(l)}{}{\mathbf{d}}$ with three different weight matrices $\mathbf{W}_{q}$, $\mathbf{W}_{k}$, and $\mathbf{W}_{v}$. This linear projection for all descriptors is expressed as

\begin{equation}
\begin{aligned}
\begin{bmatrix}
\mathbf{q} \\
\mathbf{k} \\
\mathbf{v} 
\end{bmatrix} = \begin{bmatrix}
\mathbf{W}_{q} \\
\mathbf{W}_{k} \\
\mathbf{W}_{v} 
\end{bmatrix} \prescript{(l)}{}{\mathbf{d}} + \begin{bmatrix}
\mathbf{b}_{q} \\
\mathbf{b}_{k} \\
\mathbf{b}_{v}
\end{bmatrix}.
\end{aligned}
\end{equation}

Note that the weight parameters are different in each graph layer. In practice, we also applied multi-head attention \cite{vaswani2017attention} to improve the expressivity from different representation subspaces of different weight parameters. Fig. \ref{fig_attention_1} presents example results of attention on robust descriptors.

\subsection{Updating D2S with Unlabeled Observations} \label{unlabel_learning}
\begin{algorithm}[t]
  
  \KwInput{$\mathcal{T} $, $\mathcal{D}_{\mathcal{T}}$, $\mathcal{W}_{\mathcal{T}}$, $\mathcal{U}$\\
  \begin{itemize}
    \item[-] $\mathcal{T} = \{\mathbf{I}^{i}_{\mathcal{T}}\}_{i=1}^{n}$ training images,
    \item[-] $\mathcal{D}_{\mathcal{T}}=\{\mathbf{D}^{i}_{\mathcal{T}}\}_{i=1}^{n}$ training descriptors,
    \item[-] $\mathcal{W}_{\mathcal{T}} = \{\mathbf{W}^{i}_{\mathcal{T}}\}_{i=1}^{n}$ training scene coordinates, 
    \item[-] $\mathcal{U}=\{\mathbf{I}^{i}_{\mathcal{U}}\}_{i=1}^{m}$ unlabeled images,
\end{itemize}
  }
  \KwOutput{ $ \mathcal{A}=\{\mathbf{D}^{i}_{\mathcal{U}}, \mathbf{W}^{i}_{\mathcal{U}}\}_{i=1}^{v}$
    \begin{itemize}
    \item[-] $\mathcal{D}_{\mathcal{U}}=\{\mathbf{D}^{i}_{\mathcal{U}}\}_{i=1}^{m}$ unlabeled descriptors, 
    \item[-] $\mathcal{W}_{\mathcal{U}} = \{\mathbf{W}^{i}_{\mathcal{U}}\}_{i=1}^{m}$ pseudo scene coordinates,
    \item[-] $ v \leq m$.
\end{itemize}}
    $\mathcal{D}_{\mathcal{U}} \leftarrow$ ExtractDescs$(\mathcal{U});$ \\
    $i:=1; \mathcal{A} \leftarrow \{\};$ \\
    \Do{$i \leq m, i++$}{
      $\mathbf{ids}_{\mathcal{T}} \leftarrow$ NearestVLAD$(\mathbf{I}_{\mathcal{U}}^{i}, \mathcal{T});$ \# nearest indices. \\
      $\mathcal{M} \leftarrow$ Matching$(\mathbf{D}_{\mathcal{U}}^{i}, \mathcal{D}_{\mathcal{T}}^{ \mathbf{ids}_{\mathcal{T}}});$ \\
      $s, \mathbf{W}^{i}_{\mathcal{U}} \leftarrow $ Copy$(\mathcal{M}, \mathcal{W}_{\mathcal{T}}^{\mathbf{ids}_{\mathcal{T}}});$ \\ 
      \If{$s \geq 50$}{
        $\mathcal{A} \leftarrow \mathcal{A} \cup \{\mathbf{D}^{i}_{\mathcal{U}}, \mathbf{W}^{i}_{\mathcal{U}}\};$ 
        }
        
    }
\caption{Pseudo-labelling for self-supervision}
\label{algo1}
\end{algorithm}

Due to the significant changes that can occur in environments over time since map creation, the ability to update from unlabeled data is crucial for most localization systems. We thus introduce a simple algorithm to update the proposed D2S with additional unlabeled data in a self-supervised learning manner.

Given a set of training images $\mathcal{T}$ and its labeled database of correspondent 2D--3D descriptors $\mathcal{D}_{\mathcal{T}}$ and $\mathcal{W}_{\mathcal{T}}$, we aim to find a set of pseudo-labels for new observation images $\mathcal{U}$. The pseudo-labels denote the newly generated 2D--3D correspondences of the unlabeled data $\mathcal{U}$. In detail, we extract the local descriptors of all images in $\mathcal{U}$, resulting in $\mathcal{D}_{\mathcal{U}}$. For each unlabeled image $\mathbf{I}_{\mathcal{U}}^{i}$, we then retrieve 10 nearest training images in the training database $\mathcal{T}$. Subsequently, we match two descriptor sources using a specific matching algorithm such as SuperGlue \cite{sarlin2020superglue}. Finally, we merge the top 10 matching pairs by copying their ground truth world coordinates from $\mathcal{T}$ to $\mathcal{U}$. This step is shown as line 6 of the algorithm \ref{algo1}, where the $\mathbf{W}^{i}_{\mathcal{U}}$ is the obtained pseudo scene coordinates of descriptor matrix $\mathbf{D}^{i}_{\mathcal{U}}$ and $s$ is the number of valid world coordinates. This process is summarized in algorithm \ref{algo1}.

\subsection{Loss Function} \label{loss_function}
According to Eq. \ref{ori_equation}, the output of sparse scene coordinates $\mathbf{w}$, predicted by D2S, includes four variables $(x,y,z,p)^{T}$. This output is rewritten as $(\hat{\mathbf{y}},p)^{T}$, where $\hat{\mathbf{y}}$ is the abbreviation of the estimated world coordinates vector and $p$ is the reliability probability. The loss function to minimize the estimated scene coordinates can be defined as 
\begin{equation}
    \mathcal{L}_{m} = \frac{1}{N}\sum^{N}\sum_{i=1}^{K} z_{i}\lVert \mathbf{y}_{i}-\hat{\mathbf{y}}_{i} \lVert^{2}_{2},
\label{loss_m}
\end{equation}

where $N$ is the number of mini-batch sizes, $K$ is the number of descriptors in the current frame, and $\mathbf{y}_{i}$ is the ground truth scene coordinate. We multiply the difference between estimated and ground truth world coordinates by $z_{i} \in \{0,1\}$, which is the ground truth reliability of descriptors. The goal is to ensure that the optimization focuses on robust descriptors while learning to ignore unreliable descriptors. 

Because the loss Eq. \ref{loss_m} is used to optimize only the robust features, we propose an additional loss function to simultaneously learn to be aware of this assumption, as follows:

\begin{equation}
    \mathcal{L}_{u} = \frac{1}{N}\sum^{N}\sum_{i=1}^{K}  \lVert z_{i}- \frac{1}{1+|\beta p|} \lVert^{2}_{2}.
\end{equation}

As described in Section \ref{simple_function}, $\beta$ is a scale factor chosen to make the expected reliability prediction easy to reach a small value when the input descriptors belong to high uncertainty regions.

In addition, we also apply the re-projection error loss for each 2D descriptor's position. The loss function aims to align the general scene coordinates following the correct camera rays, which is described as follows:

\begin{equation}
    \mathcal{L}_{r} = \frac{1}{N}\sum_{j=1}^{N}\sum_{i=1}^{K}z_{i}\lVert \pi(\mathbf{R}_{j} \mathbf{y}_{i}+\mathbf{t}_{j}) - \mathbf{u}_{i}\lVert^{2}_{2},
    \label{reproject_loss}
\end{equation} 

where $\mathbf{R}_{j}$ is the rotation matrix and $\mathbf{t}_{j}$ is the translation vector of camera pose $j$ obtained from SfM models. $\pi(.)$ is a function to convert the estimated 3D camera coordinates to 2D pixel position, with $u_{i}$ as its ground truth pixel coordinate. $z_{i} \in \{0,1\}$, where $z_i = 1$ if keypoint has 3D coordinate available in SfM models, and $z_i = 0$ otherwise.

Finally, these loss functions are combined with the scale factor as follows:

\begin{equation}
    \mathcal{L}=\alpha_{m}\mathcal{L}_{m}+\alpha_{u}\mathcal{L}_{u}+\alpha_{r}\tau(t) tanh\bigg(\frac{\mathcal{L}_{r}}{\tau(t)}\bigg),
    \label{total_loss}
\end{equation}
where $\tau(t) = \omega(t) \tau_{max} + \tau_{min},  \text{ with } \omega(t) = \sqrt{1-t^{2}}$, $t\in(0,1)$ denotes the relative training progress \cite{brachmann2023accelerated}. The $\alpha_{r} = 1$ if $\frac{c_{i}}{c_{total}}>0.8$, otherwise $\alpha_{r} = 0$, where $c_{i}$ and $c_{total}$ are current and total training iterations respectively. 
\subsection{Camera Pose Estimation} \label{pose_estimation}
\begin{figure}[t]
    \centering
    \hspace*{-0.1cm}
    \includegraphics[width = 0.5\textwidth]{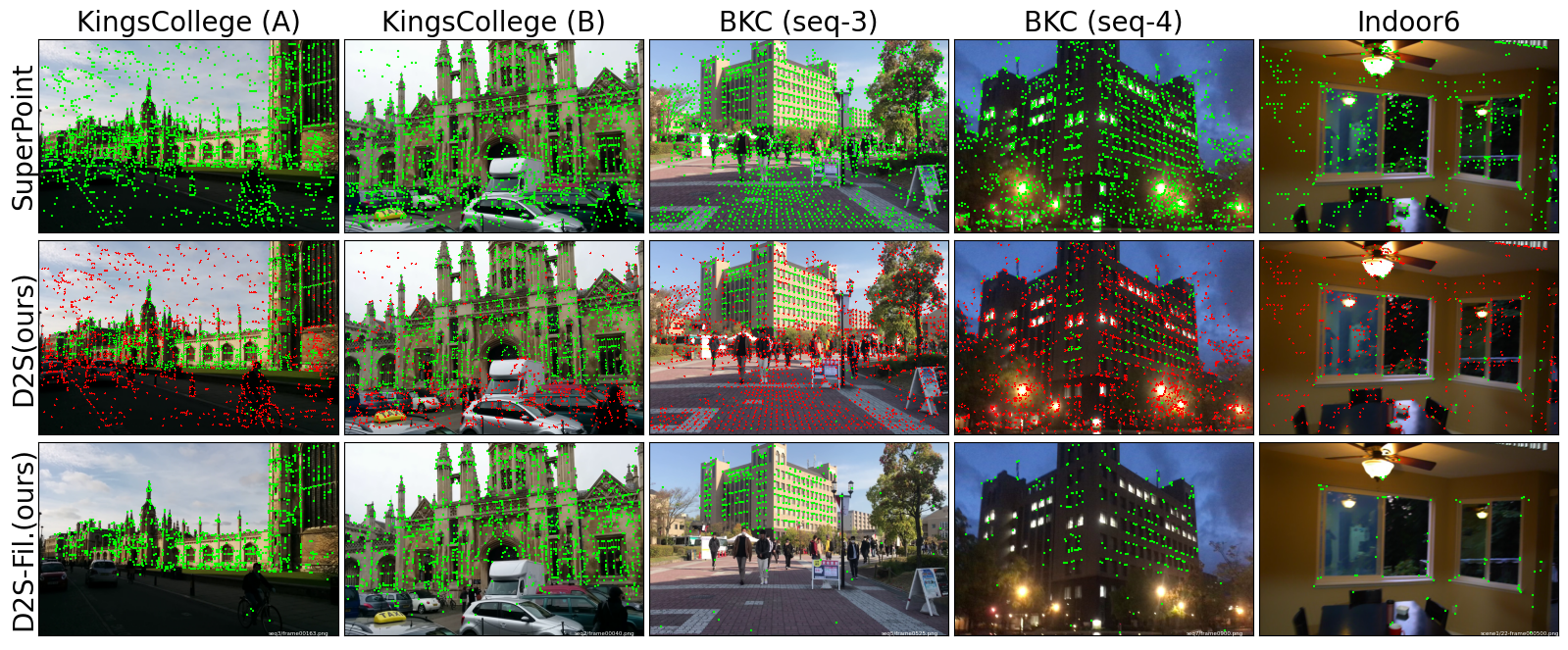}
    \caption{\textbf{Reliability Prediction Results}. The \textcolor{red}{red} points denote the low predicted reliability for localization, whereas \textcolor{green}{green} points are the predicted robust ones.}
    \label{Uncertainties_results}
\end{figure}

We estimate the camera pose using the most robust 3D coordinates based on their predicted reliability $\hat{z}$. We visualize examples of predicted robust coordinates in Fig. \ref{Uncertainties_results}. Finally, a robust minimal solver (PnP+RANSAC\cite{PoseLib}) is used, followed by a Levenberg--Marquardt-based nonlinear refinement, to compute the camera pose.

In contrast to DSAC$^{*}$ \cite{brachmann2021visual} and several SCR-based works \cite{brachmann2017dsac, brachmann2018learning, li2020hierarchical}, which only focus on estimating dense 3D scene coordinates of the input image, our D2S learn to detect of salient keypoints and its coordinates. We believe that the SCR network should learn to pay more attention to salient features; this can lead to a better localization performance \cite{do2022learning}. This is also supported by empirical evidence of D2S's results.



\section{Experiments} \label{experiments}

We evaluate the proposed re-localization method on three different datasets: one indoor\cite{do2022learning} and two outdoor datasets\cite{kendall2015posenet} including a proposed dataset. We set $L=5$ for the number of graph attention with a subsequent MLP of $(D, 512, 1024, 1024, 512, 4)$, where $D=256$ is the dimensional number of SuperPoint\cite{detone2018superpoint} features. We used Pytorch \cite{paszke2019pytorch} to implement the proposed method and trained the network with 1.5M iterations with the batch size of a single image. The learning rate was set at $0.0001$ with the Adam optimizer \cite{kingma2014adam}, being reduced with a decay of 0.5 for every one-seventh total iterations. For data augmentation, we randomly applied brightness and contrast change of $\pm15\%$ and rotated and resized the image $\pm 30^{\circ}$. The random rate for applying augmentation on an image was 50\% in each training batch. All the experiments were evaluated using an Nvidia GeForce GTX 1080ti GPU and an Intel Core i9-9900. The training time for each scene took approximately 9.4 hours.

\subsection{Localization Results on Indoor-6} 
\label{sec_datasets}
The Indoor-6 dataset \cite{do2022learning}, composed of images captured at different times and under varying lighting conditions, poses a considerable challenge. Table \ref{Indoor-6_results} shows the results of our method. D2S consistently surpasses PoseNet \cite{kendall2015posenet}, DSAC* \cite{brachmann2021visual}, ACE \cite{brachmann2023accelerated}, and NBE+SLD \cite{do2022learning} in localization accuracy across all six scenes.

Specifically, D2S achieves a 28.4\% improvement in Scene 2a compared to NBE+SLD, lowering the error from 7.2cm/0.68° to 4.0cm/0.41°. Averaged across the six scenes, D2S records an error of 3.38cm/0.62° and an improvement of 16.2\% in accuracy.

In terms of storage, NBE+SLD requires two separate models, consuming 132 MB, while our method reduces this to a single network requiring only 22 MB, achieving six times the storage efficiency. Although ACE \cite{brachmann2023accelerated} uses the least memory, its performance is 28.3\% lower than that of D2S, highlighting the robustness of our approach in diverse environmental conditions.

\begin{table*}
\centering
\caption{\textbf{Localization results on Indoor-6.} We report the median errors in cm for the position, degree ($^{\circ}$) for the orientation, and accuracy at 5cm/5$^{\circ}$ of the Indoor-6 dataset. The size column indicates storage requirements for each method. The best regression-based results are \textbf{in bold} and the second best is in \underline{underlined}.}
\label{Indoor-6_results}
\resizebox{\linewidth}{!}{%
\begin{tabular}{c|c|cc|cc|cc|cc|cc|cc|cc} 
\hline
\multicolumn{16}{c}{Indoor-6~~}                                                                                                                                                                                                                                                                     \\ 
\hline
\multirow{2}{*}{Method} & Size       & \multicolumn{2}{c|}{scene1}       & \multicolumn{2}{c|}{scene2a}      & \multicolumn{2}{c|}{scene3}       & \multicolumn{2}{c|}{scene4a}      & \multicolumn{2}{c|}{scene5}       & \multicolumn{2}{c|}{scene6}       & \multicolumn{2}{c}{Average}         \\
                        & (MB)       & (cm/deg.)         & (\%)          & (cm/deg.)         & (\%)          & (cm/deg.)         & (\%)          & (cm/deg.)         & (\%)          & (cm/deg.)         & (\%.)         & (cm/deg.)         & (\%)          & (cm/deg.)          & (\%)           \\ 
\hline
Hloc$^{\text{FM-based}}$ \cite{sarlin2019coarse, sarlin2020superglue}                   & 730–2360   & 3.2/0.47          & 64.8          & -/-               & 51.4          & 2.1/0.37          & 81.0          & -/-               & 69.0          & 6.1/0.86          & 42.7          & 2.1/0.42~         & 79.9          & -/-                & 64.8           \\ 
\hline
PoseNet \cite{kendall2015posenet}                & 50         & 159.0/7.46        & 0.0           & -/-               & -             & 141.0/9.26        & 0.0           & -/-               & -             & 179.3/9.37        & 0.0           & 118.2/9.26        & 0.0           & -/-                & -              \\
DSAC* \cite{brachmann2021visual}                  & 28         & 12.3/2.06         & 18.7          & 7.9/0.9           & 28.0          & 13.1/2.34         & 19.7          & 3.7/0.95          & 60.8          & 40.7/6.72         & 10.6          & 6.0/1.40          & 44.3          & 13.9/2.39          & 30.4           \\
ACE \cite{brachmann2023accelerated}                    & \textbf{4} & 13.6/2.1          & 24.9          & 6.8/0.7           & 31.9          & 8.1/1.3           & 33.0          & 4.8/0.9           & 55.7          & 14.7/2.3          & 17.9          & 6.1/1.1           & 45.5          & 9.02/1.40          & 34.8           \\
NBE+SLD(E) \cite{do2022learning}             & 29         & 7.5/1.15          & 28.4          & 7.3/0.7           & 30.4          & 6.2/1.28          & 43.5          & 4.6/1.01          & 54.4          & 6.3/0.96          & 37.5          & 5.8/1.3           & 44.6          & 6.28/1.07          & 39.8           \\
NBE+SLD \cite{do2022learning}                & 132        & \underline{6.5/0.90}  & \underline{38.4}  & \underline{7.2/0.68}  & \underline{32.7}  & \underline{4.4/0.91}  & \underline{53.0}  & \underline{3.8/0.94}  & \underline{66.5}  & \underline{6.0/0.91}  & \underline{40.0}  & \underline{5.0/0.99}  & \underline{50.5}  & \underline{5.48/0.89}  & \underline{46.9}   \\
D2S (\textbf{ours})     & \underline{22} & \textbf{4.8/0.81} & \textbf{51.8} & \textbf{4.0/0.41} & \textbf{61.1} & \textbf{3.6/0.69} & \textbf{60.0} & \textbf{2.1/0.48} & \textbf{84.8} & \textbf{5.8/0.90} & \textbf{45.5} & \textbf{2.4/0.48} & \textbf{75.2} & \textbf{3.38/0.62} & \textbf{63.1}  \\
\hline
\end{tabular}
}
\end{table*}

\begin{table*}[t]
\centering
\caption{\textbf{Localization results on Cambridge Landmarks.}}
\label{all_results_cambridge}
\resizebox{0.88\linewidth}{!}{%
\begin{tabular}{c|c|cc|cc|cc|cc|cc|cc} 
\hline
\multicolumn{14}{c}{Cambridge Landmarks}                                                                                                                                                                                                                                                                                                                                                      \\ 
\hline
\multicolumn{2}{c|}{\multirow{3}{*}{Method}} & \multicolumn{2}{c|}{King's College} & \multicolumn{2}{c|}{Old Hospital} & \multicolumn{2}{c|}{Shop Facade} & \multicolumn{2}{c|}{St Mary's Church}                      & \multicolumn{2}{c|}{Great Court}                                                                                                  & \multicolumn{2}{c}{Average}      \\ 
\cline{3-14}
\multicolumn{2}{c|}{}                        & Size       & Error                  & Size       & Error                & Size       & Error               & Size       & Error                                         & Size       & Error                                                                                                                & Size       & Error               \\
\multicolumn{2}{c|}{}                        & (MB)       & (m/deg.)               & (MB)       & (m/deg.)             & (MB)       & (m/deg.)            & (MB)       & (m/deg.)                                      & (MB)       & (m/deg.)                                                                                                             & (MB)       & (m/deg.)            \\ 
\hline
\multirow{3}{*}{FM}  & Hloc \cite{sarlin2019coarse, sarlin2020superglue}                 & 1877       & 0.06/0.10              & 1335       & 0.13/0.23            & 316        & 0.03/0.14           & 2009       & 0.04/\textcolor[rgb]{0.122,0.122,0.157}{0.13} & 1746       & \textcolor[rgb]{0.122,0.122,0.157}{0.1}\textcolor[rgb]{0.122,0.122,0.157}{/}\textcolor[rgb]{0.122,0.122,0.157}{0.05} & 1459       & 0.07/0.13           \\
                     & AS \cite{sattler2016efficient}                   & 275        & 0.57/0.70              & 140        & 0.52/1.12            & 37.7       & 0.12/0.41           & 359        & 0.22/0.62                                     & -          & \textcolor[rgb]{0.122,0.122,0.157}{0.24/0.13}                                                                        & -          & 0.33/0.60           \\
                     & SS \cite{yang2022scenesqueezer}                   & 0.3        & 0.27/0.38              & 0.53       & 0.37/0.53            & 0.13       & 0.11/0.38           & 0.95       & 0.15/0.37                                     & -          & -/-                                                                                                                  & -          & -/-                 \\ 
\hline
\multirow{3}{*}{APR} & PoseNet \cite{kendall2015posenet}              & 50         & 0.88/1.04              & 50         & 3.2/3.29             & 50         & 0.88/3.78           & 50         & 1.6/3.32                                      & 50         & 6.83/3.47                                                                                                            & 50         & 2.78/2.98           \\
                     & FeatLoc \cite{bach2022featloc}              & 34         & 1.3/3.84               & 34         & 2.1/6.1              & 34         & 0.91/7.5            & 34         & 3.0/10.4                                      & -          & -/-                                                                                                                  & -          & -/-                 \\
                     & DFNet \cite{chen2022dfnet}                & 60         & 0.73/2.37              & 60         & 2.0/2.98             & 60         & 0.67/2.21           & 60         & 1.37/4.03                                     & -          & -/-                                                                                                                  & -          & -/-                 \\ 
\hline
\multirow{6}{*}{SCR} & DSAC++ \cite{brachmann2018learning}               & 104        & 0.18/\underline{0.3}       & 104        & 0.20/\underline{0.3}     & 104        & 0.06/0.3            & 104        & 0.13/0.4                                      & 104        & 0.40/0.2                                                                                                             & 104        & 0.19/ 0.30          \\
                     & SCoCR \cite{li2020hierarchical}                & 165        & 0.18/\underline{0.3}       & 165        & \underline{0.19/0.3}     & 165        & 0.06/0.3            & 165        & \underline{0.09}/0.3                              & 165        & \underline{0.28/0.2}                                                                                                     & 165        & \underline{0.16/0.28}   \\
                     & DSAC* \cite{brachmann2021visual}                & 28         & \underline{0.15/0.3}       & 28         & 0.21/0.4             & 28         & \underline{0.05}/0.3    & 28         & 0.13/0.4                                      & 28         & \textcolor[rgb]{0.122,0.122,0.157}{0.49/0.3}                                                                         & 28         & 0.21/0.34           \\
                     & ACE \cite{brachmann2023accelerated}                  & \textbf{4} & 0.28/0.4               & \textbf{4} & 0.31/0.6             & \textbf{4} & \underline{0.05}/0.3    & \textbf{4} & 0.18/0.6                                      & \textbf{4} & 0.43/\underline{0.2}                                                                                                     & \textbf{4} & 0.25/0.42           \\
                     & HSCNet++ \cite{wang2024hscnet++}              & 85         & 0.19/0.34              & 85         & 0.20/0.31            & 85         & 0.06/\underline{0.24}   & 85         & \underline{0.09/0.27}                             & 85         & 0.39/0.23                                                                                                            & 85         & 0.19/\underline{0.28}   \\
                     & D2S (\textbf{ours})   & \underline{22} & \textbf{0.07/0.12}     & \underline{22} & \textbf{0.15/0.29}   & \underline{22} & \textbf{0.03/0.17}  & \underline{22} & \textbf{0.08/0.25}                            & \underline{22} & \textbf{0.23/0.11}                                                                                                   & \underline{22} & \textbf{0.11/0.19}  \\
\hline
\end{tabular}
}
\end{table*}

\subsection{Localization Results on Cambridge Landmarks}
\label{outdoor_results_cambridge}
To evaluate the quality of re-localization performance on the Cambridge Landmarks\cite{kendall2015posenet} dataset, we report the median pose error for each scene, and the results are displayed in Table \ref{all_results_cambridge}. In comparison, APR-based approaches including PoseNet \cite{kendall2015posenet}, FeatLoc \cite{bach2022featloc}, and DFNet \cite{chen2022dfnet}, achieved the lowest performance with low memory demand for all scenes. In contrast, the FM-based Hloc \cite{sarlin2020superglue, sarlin2019coarse} pipeline achieved the highest re-localization accuracy but requires a high memory footprint. We also report the results of the compressed version  SceneSqueezer (SS) \cite{yang2022scenesqueezer} on this dataset, which has a significantly low memory footprint. However, its accuracy is much worse compared to Hloc.

Importantly, we report the results of SRC-based approaches as well as the storage demand of each scene. The proposed D2S achieved the highest performance in all scenes. In detail, for example, in the King's College scene, our method shows 53\% position improvement compared to the second-best method of DSAC*\cite{brachmann2021visual}. The reduction of error is from 0.15m/0.3$^{\circ}$ to 0.07m/0.12$^{\circ}$. On average, our method marks the top-1 accuracy among learning-based localization methods, which improved by 31.1\% in positional error compared to the second-best method of SCoCR \cite{li2020hierarchical}.

\begin{figure}
    \centering
    \includegraphics[width=0.5\textwidth]{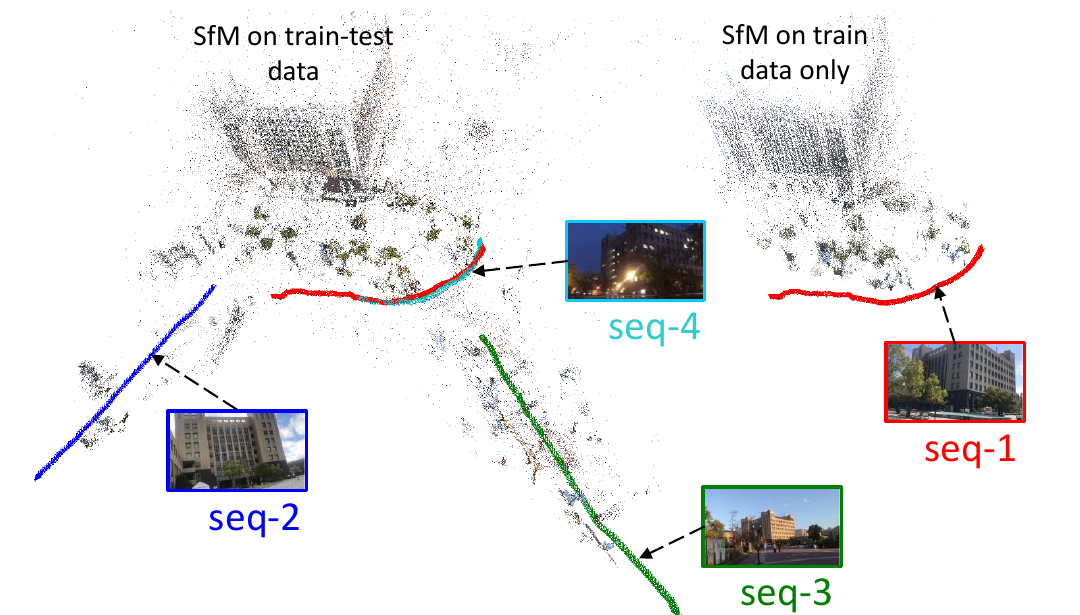}
    \caption{\textbf{Proposed challenge Ritsumeikan BKC dataset}. The dataset consists of different image sequences used for \textcolor{red}{training} (daytime), testing (indicates in \textcolor{blue}{blue}, \textcolor{green}{green}-high domain shifts, and \textcolor{cyan}{cyan}-nighttime), and unlabeled sequences for self-supervision.}
    \label{BKC_train_test}
\end{figure}

\begin{table}
\centering

\caption{\textbf{Results on Ritsumeikan BKC.} We report the median errors in comparison with other methods on \textit{BKC} dataset. The accuracy is calculated at 0.5m/10$^{\circ}$. The methods indicated with $^{\dagger}$ are self-update with unlabeled data.}
\resizebox{\linewidth}{!}{%
\begin{tabular}{c|cc|cc|cc|cc} 
\hline
\multirow{5}{*}{Method}       & \multicolumn{8}{c}{Ritsumeikan BKC}                                                                                                                                                                                                                    \\ 
\cline{2-9}
                              & \multicolumn{2}{c|}{Seq-2~}                                                                                                 & \multicolumn{2}{c|}{Seq-3~}         & \multicolumn{2}{c|}{Seq-4~~}       & \multicolumn{2}{c}{\multirow{2}{*}{Average}}  \\
                              & \multicolumn{2}{c|}{(domain-shift)}                                                                                         & \multicolumn{2}{c|}{(domain-shift)} & \multicolumn{2}{c|}{(nighttime)}   & \multicolumn{2}{c}{}                          \\ 
\cline{2-9}
                              & Errors                                                                                             & Acc.                   & Errors             & Acc.           & Errors             & Acc.          & Errors             & Acc.                     \\
                              & (m/deg.)                                                                                           & (\%)                   & (m/deg.)           & (\%)           & (m/deg.)           & (\%)          & (m/deg.)           & (\%)                     \\ 
\hline
Hloc \cite{sarlin2019coarse, sarlin2020superglue}                         & 0.25/2.12                                                                                          & 57.3                   & 0.04/0.19          & 83.5           & 0.03/0.30          & 74.1          & 0.11/0.87          & 71.6                     \\ 
\hline
DSAC* \cite{brachmann2021visual}                        & 29.6/67.1                                                                                          & 0.0                    & 8.92/28.5          & 0.0            & 8.85/57.0          & 0.0           & 15.8/50.8          & 0.0                      \\
ACE \cite{brachmann2023accelerated}                          & 0.18/1.7                                                                                           & 53.4                   & 8.64/60.2          & 18.3           & 3.89/59.5          & 2.5           & 4.24/40.5          & 24.7                     \\
$^\dagger$ACE0 \cite{brachmann2024scene}               & \textbf{\textbf{\textcolor[rgb]{0.122,0.133,0.157}{0.11/}\textcolor[rgb]{0.122,0.133,0.157}{1.0}}} & \textbf{\textbf{88.3}} & 6.56/54.9          & 17.4           & 5.09/74.8          & 1.2           & 3.92/43.6          & 35.6                     \\
D2S(\textbf{\textbf{ours}})   & 0.43/3.62                                                                                          & 51.5                   & \underline{0.36/2.24}  & \underline{56.0}   & \underline{0.18/2.12}  & \underline{63.0}  & \underline{0.32/2.6}   & \underline{56.8}             \\
$^\dagger$D2S+(\textbf{ours}) & \underline{0.16/1.34}                                                                                  & \underline{75.7}           & \textbf{0.09/0.48} & \textbf{93.6}  & \textbf{0.08/1.29} & \textbf{70.4} & \textbf{0.11/1.04} & \textbf{79.9}            \\
\hline
\end{tabular}
}
\label{BKC_median_results}
\end{table}

\subsection{Results for Outdoor Localization on BKC Dataset} \label{outdoor_results_BKC}

This section reports the performance of the proposed method and other baselines on the proposed Ritsumeikan BKC dataset, illustrated in Fig. \ref{BKC_train_test}. This dataset has not been evaluated by previous localization methods; therefore, we selected three state-of-the-art baselines, FM-based Hloc \cite{sarlin2019coarse, sarlin2020superglue}, SCR-based DSAC* \cite{brachmann2021visual} and ACE\cite{brachmann2023accelerated}, to draw a comparison with the proposed D2S. We used the same training and localization configurations stated in the original papers\cite{sarlin2019coarse, brachmann2021visual,brachmann2023accelerated}. 

Table \ref{BKC_median_results} reports the median localization errors of ACE, DSAC*, Hloc, and the proposed D2S on different test sequences of the BKC dataset. The BKC dataset features several challenges for visual localization methods on their capacity to generalize beyond the training data (high domain shifts and transition from day to night). Thus, it is particularly challenging for SCR approaches because significant domain shifts can result in substantial differences between training and test images. This is evident in the DSAC* and ACE's worse results on testing sequences. On the other hand, the proposed D2S outperforms DSAC* and ACE with a remarkable improvement in localization accuracy of 32.1\%.

\textbf{Self-update with unlabeled data (D2S+).} Importantly, the BKC dataset is proposed to evaluate localization methods in the capability of updating with unlabeled observations. Note that the previous datasets \cite{ kendall2015posenet, do2022learning} do not offer any available new observations for evaluating self-supervision. In the BKC dataset, each test sequence in has a corresponding unlabeled sequence recorded at different times. For self-supervision, we update the model with an additional 50k iterations using data generated by Algorithm \ref{algo1}.


We also report the results of D2S+ when updated with unlabeled data in Table \ref{BKC_median_results}. Notably, the improvement in accuracy obtained through D2S, is from 56.8\% to 79.9\%, within the threshold of 0.5m/10$^{\circ}$. D2S+ even surpasses Hloc and ACE0 \cite{brachmann2024scene} accuracy by 8.3\% and 44.3\% respectively.



\begin{figure}
    \centering
    \includegraphics[width=\linewidth]{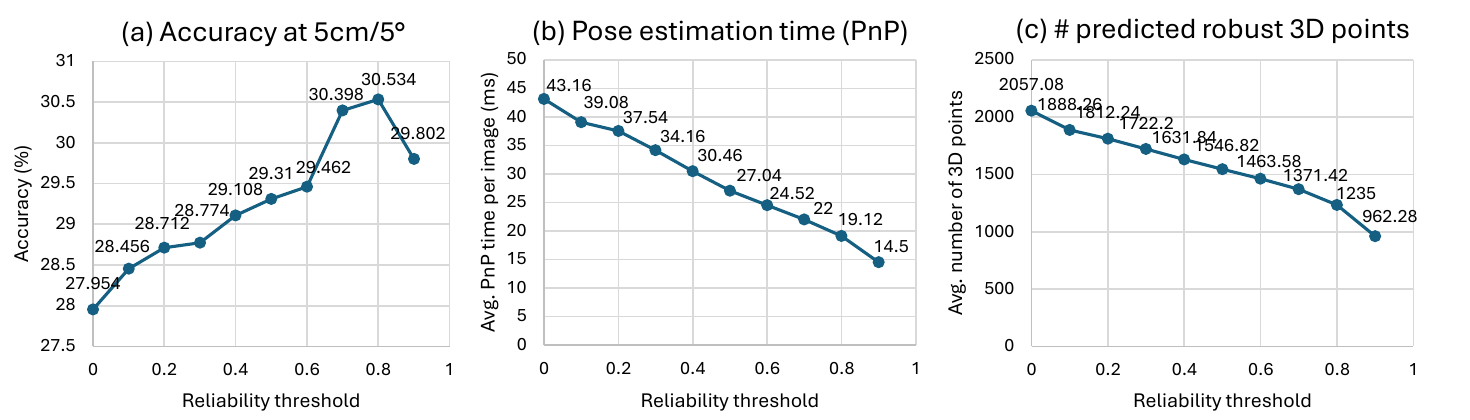}
    \caption{\textbf{Reliability Effects}. We report the impact of reliability prediction on re-localization performance using the Cambridge Landmarks dataset.}
    \label{reliability_changing}
\end{figure}

\begin{table}[]
    \centering
    \caption{Results when using different feature extractors.}
    \resizebox{0.9\linewidth}{!}{%
    \begin{tabular}{c|c|c|cc} 
\hline
\multirow{3}{*}{\begin{tabular}[c]{@{}c@{}}Localization\\~Method\end{tabular}} & \multirow{3}{*}{Extractor} & \multirow{3}{*}{\begin{tabular}[c]{@{}c@{}}Desc.~\\dim.\end{tabular}} & \multirow{2}{*}{Size} & Cambridge~          \\
                                                                               &                            &                                                                       &                       & KingsCollege        \\
                                                                               &                            &                                                                       & (MB)                  & (m / deg.)          \\ 
\hline
\multirow{3}{*}{Hloc \cite{sarlin2019coarse, sarlin2020superglue}}                                                          & SIFT \cite{lowe2004distinctive}~                      & 128                                                                   & 849                   & 0.05/0.09           \\
                                                                               & R2D2 \cite{revaud2019r2d2}                       & 128                                                                   & 699                   & 0.05/0.09           \\
                                                                               & SuperPoint \cite{detone2018superpoint}                 & 256                                                                   & 1877                  & 0.06/0.10           \\ 
\hline
\multirow{3}{*}{D2S (ours)}                                                    & SIFT \cite{lowe2004distinctive}                      & 128                                                                   & \textbf{12}           & 0.15/0.19           \\
                                                                               & R2D2 \cite{revaud2019r2d2}                       & 128                                                                   & \textbf{12}           & 0.08/\textbf{0.12}  \\
                                                                               & SuperPoint \cite{detone2018superpoint}                 & 256                                                                   & 22                    & \textbf{0.07/0.12}  \\
\hline
\end{tabular}
}
    \label{differnt_feature_extractor}
\end{table}

\begin{figure}[t]
    \centering
    \hspace*{-0.2cm}
    \includegraphics[width=0.46\textwidth]{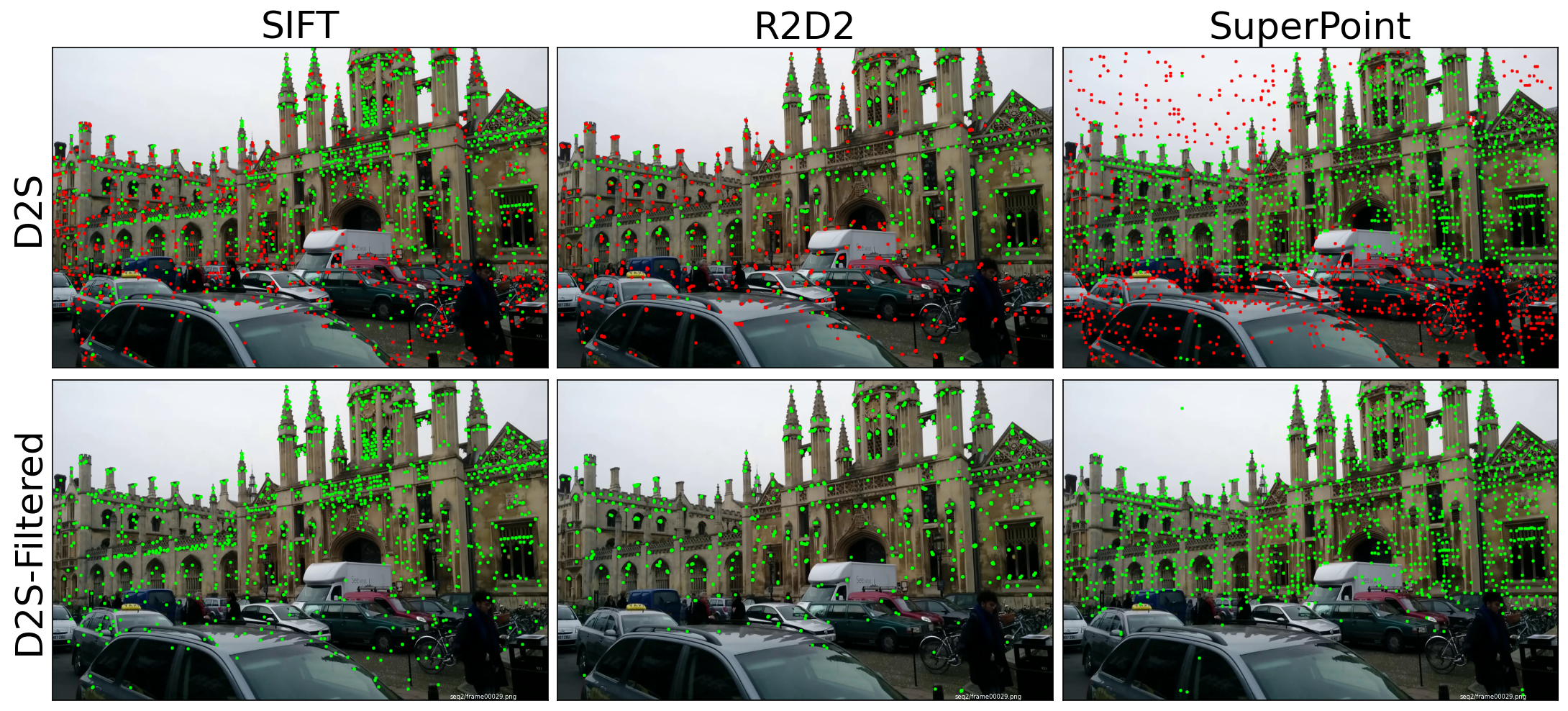}
    \caption{Predicted uncertainty results using different descriptors including SIFT\cite{lowe2004distinctive}, R2D2\cite{revaud2019r2d2}, and SuperPoint\cite{detone2018superpoint}. }
    \label{uncertainty_ablation}
\end{figure}

\begin{figure}
    \centering
    \includegraphics[width=\linewidth]{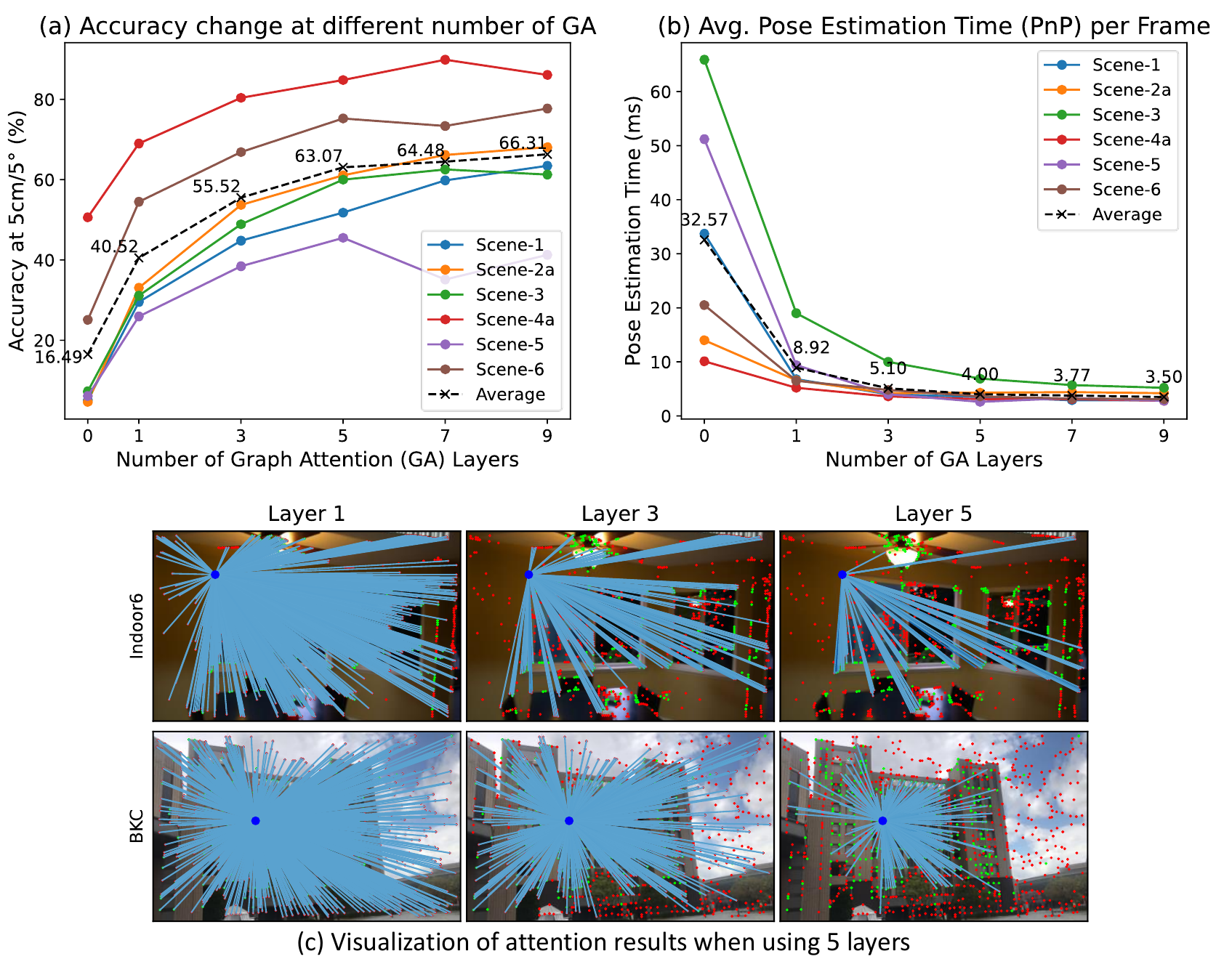}
    \caption{Ablation study in changing the number of graph attention layers in Indoor-6.}
    \label{changing_graph_attention_Indoor-6}
\end{figure}

\subsection{Ablation Study}
\textbf{System efficiency.}
 The proposed method achieves an average processing speed of 21 frames per second (FPS) on the Cambridge dataset and 55 FPS on the Indoor-6 dataset. When compared to NBE+SLD \cite{do2022learning} on the Indoor-6 dataset, D2S demonstrates a significant advantage, being 4.8X  faster in inference speed and even more accurate (63.1\% .vs 46.9\%). NBE+SLD requires approximately 132MB, whereas the proposed method is 6X smaller by using only 22 MB for storing D2S parameters.
In the details of the D2S pipeline, the SuperPoint feedforward process takes 10 ms for an Indoor-6 image with a resolution of 640 pixels, and 19 ms for a Cambridge image with a resolution of 1024 pixels. Additionally, the computation of 3D coordinates in D2S requires 4.0 ms. The PnP RANSAC process takes an average of 24.5 ms on the Cambridge dataset and 4.0 ms on the Indoor-6 dataset.

\textbf{Reliability effects.} The proposed method effectively eliminates low-reliability descriptors, achieved through the use of the simple loss function presented in Equation \ref{reliable_equation} in Section \ref{simple_function}. We guide this loss function using the pre-built SfM model for each scene. An ablation study demonstrating this capability is provided in Fig. \ref{reliability_changing}, which shows a significant improvement in localization performance upon removing the uncertain descriptors identified by D2S.

\textbf{Different feature extractors (FEs).} We present a brief comparison between the proposed D2S and FM-based Hloc when using different FEs, as shown in Table \ref{differnt_feature_extractor} and Fig. \ref{uncertainty_ablation}. The results demonstrate that the proposed D2S can significantly reduce memory requirements for localization using different FEs while maintaining comparable accuracies.

\textbf{Graph Attention.} We show that graph attention (GA) layers are critical for achieving the best performance of D2S, shown in Fig. \ref{changing_graph_attention_Indoor-6}. We evaluate our method by varying the number of GA layers from 0 to 9 on the Indoor-6 dataset. As depicted in Fig. \ref{changing_graph_attention_Indoor-6}(a), without GA layers, D2S reaches only 16.5\% accuracy, while the introduction of a single GA layer raises accuracy to 40.52\%. The highest accuracy of 66.3\% is observed with 9 GA layers. However, to ensure a lightweight model, we utilize only 5 GA layers in this study. Additionally, Fig. \ref{changing_graph_attention_Indoor-6}(b) shows that the inclusion of GA layers improves pose estimation time, likely due to the enhanced quality of 3D point predictions with more GA layers.

\textbf{Loss functions.} We highlight the importance of the loss components \(\mathcal{L}_{u}\) and \(\mathcal{L}_{r}\) in training D2S by reporting the average median errors and accuracies at thresholds of 5cm/5\% on the Cambridge Landmarks dataset. When both \(\mathcal{L}_{u}\) and \(\mathcal{L}_{r}\) are utilized, D2S achieves its best performance with an average median error of 11.2cm/0.19\(^{\circ}\), and an accuracy of 30.5\%. Dropping \(\mathcal{L}_{u}\) leads to a performance decline, increasing the average median error to 14.1cm/0.23\(^{\circ}\), while reducing the accuracy to 24.3\%. Similarly, excluding \(\mathcal{L}_{r}\) results in an average median error of 11.6cm/0.20\(^{\circ}\), with an accuracy of 25.9\%.

\textbf{Limitations.} Since the proposed method relies on sparse keypoints, it inherits the typical challenges faced by sparse feature-based methods, particularly in low-texture or repetitive-structure regions. Additionally, when applied to large-scale environments, it is recommended to train multiple D2S models for each sub-region to ensure optimal performance. However, this approach may result in significantly longer training times.

\section{Conclusion} \label{conclusion}
In this study, we introduced D2S, a novel direct sparse regression approach for establishing 2D--3D correspondences in visual localization. Our proposed pipeline is both simple and cost-effective, enabling the efficient representation of complex descriptors in 3D SfM maps. By utilizing an attention graph and straightforward loss functions, D2S was able to accurately classify and learn to focus on the most reliable descriptors, leading to significant improvements in re-localization performance. We conducted a comprehensive evaluation of our proposed D2S method on three distinct indoor and outdoor datasets, demonstrating its superior re-localization accuracy compared to state-of-the-art baselines. Furthermore, we introduced a new dataset for assessing the generalization capabilities of visual localization approaches. The obtained results demonstrate that D2S can generalize beyond its training data, even in challenging domain shifts such as day-to-night transitions and in locations lacking labeled data. Additionally, D2S is capable of being self-supervised with new observations, eliminating the need for camera poses, camera parameters, and ground truth scene coordinates.

\bibliographystyle{IEEEtran}
\bibliography{reference.bib}
\end{document}